\documentclass[10pt,twocolumn,letterpaper]{article}

\usepackage{cvpr}              

\usepackage{graphicx}
\usepackage{amsmath}
\usepackage{amssymb}
\usepackage{booktabs}

\newcommand{\argmax}{\operatornamewithlimits{arg\,max}}

\usepackage[table]{xcolor}
\usepackage{color}
\usepackage{times}
\usepackage{epsfig}
\usepackage{amsbsy}
\usepackage{xcolor}
\usepackage{xspace}
\usepackage{algorithm}
\usepackage{algorithmic}
\usepackage{multirow}
\usepackage{multicol}
\usepackage{float}
\usepackage{relsize}
\usepackage{soul}
\usepackage{tabularx}
\usepackage{makecell}

\definecolor{darkgreen}{rgb}{0,0.45,0}

\usepackage[pagebackref,breaklinks,colorlinks]{hyperref}

\usepackage[capitalize]{cleveref}
\crefname{section}{Sec.}{Secs.}
\Crefname{section}{Section}{Sections}
\Crefname{table}{Table}{Tables}
\crefname{table}{Tab.}{Tabs.}

\def\cvprPaperID{11131} 
\def\confName{CVPR}
\def\confYear{2022}

\begin{document}
\newcolumntype{L}{>{\centering\arraybackslash}m{1cm}}
\title{Contrastive Test-Time Adaptation}

\author{Dian Chen\thanks{Work done when author previously worked at UC Berkeley.}\\
Toyota Research Institute\\
\and
Dequan Wang\\
UC Berkeley\\
\and
Trevor Darrell\\
UC Berkeley\\
\and
Sayna Ebrahimi\thanks{Author is now at Google.}\\
UC Berkeley
}
\maketitle

\begin{abstract}
Test-time adaptation is a special setting of unsupervised domain adaptation where a trained model on the source domain has to adapt to the target domain without accessing source data. We propose a novel way to leverage self-supervised contrastive learning to facilitate target feature learning, along with an online pseudo labeling scheme with refinement that significantly denoises pseudo labels. The contrastive learning task is applied jointly with pseudo labeling, contrasting positive and negative pairs constructed similarly as MoCo but with source-initialized encoder, and excluding same-class negative pairs indicated by pseudo labels. Meanwhile, we produce pseudo labels online and refine them via soft voting among their nearest neighbors in the target feature space, enabled by maintaining a memory queue. Our method, AdaContrast, achieves state-of-the-art performance on major benchmarks while having several desirable properties compared to existing works, including memory efficiency, insensitivity to hyper-parameters, and better model calibration. Project page: \url{sites.google.com/view/adacontrast}. 

\end{abstract}

%

\section{Introduction}
\label{sec:intro}




Deep networks are remarkably successful in learning visual tasks when training and test data follow the same distribution. However, their ability to generalize to unseen data suffers in the presence of \textit{domain shift}~\cite{covariateshift,datasetshift}. Building models that can adapt to distribution shifts is the focus of domain adaptation where the goal is to transfer knowledge from a labeled \textit{source} domain to a new but related \textit{target} domain~\cite{saenko2010adapting, long2015learning, ganin2016domain,bousmalis2017unsupervised, tzeng2017adversarial,hoffman2018cycada}. In this work we focus on the problem of test-time~\cite{sun19ttt,wang2021tent} or source-free\cite{liang2020we,yang2021generalized,kundu2020universal} domain adaptation where the source data is no longer available during adapting to unlabeled test data. Since test-time adaptation (TTA) only requires access to the source model, 
it is appealing to real-world applications where data privacy and transmission bandwidth become critical issues.

\begin{figure}[t]
  \centering
   \includegraphics[width=\linewidth]{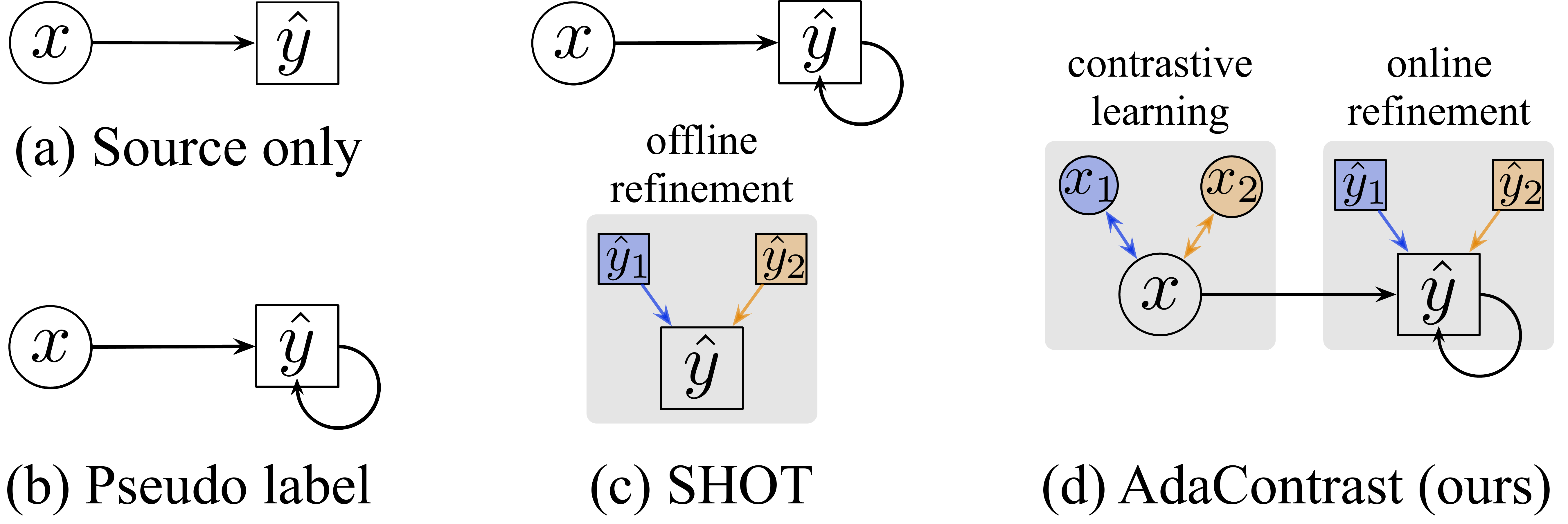}
   \caption{Illustration of how our method, AdaContrast, leverages target domain data vs. prior works. (a) Without adaptation, source-only model is only evaluated on target data. (b) With pseudo labeling, the source classifier predictions are used as pseudo labels for self-training. (c) Existing pseudo labeling approach, SHOT~\cite{liang2020we}, uses offline global refinement to reduce noisy pseudo labels. (d) In AdaContrast, we consider two kinds of relations among target samples: we use contrastive learning to exploit information from sample pairs to learn better target representation, while refining pseudo labels by aggregating knowledge in the neighborhood. Colors indicate pseudo-labeled classes.}
   \label{fig:teaser}
\end{figure}

However, the challenging setting of TTA raises two major questions: 1) how to learn the target-domain representation without the help of ground truth annotation 2) how to build the target-domain classifier with only the source-domain classifier available as a proxy for the source domain. 
To address these difficulties, existing works have leveraged image/feature generation \cite{li2020model,kundu2020universal}, class prototypes \cite{liang2020we,yang2020unsupervised}, entropy minimization \cite{liang2020we,wang2021tent}, self-training or pseudo labeling \cite{liang2020we}, and self-supervised auxiliary task training~\cite{sun19ttt}. Generative models have the drawback of requiring a large computation capacity for generating target-style images/features~\cite{li2020model}. Entropy minimization-based methods have been competitive but the direct optimization of entropy disrupts the model calibration on target. Pseudo-labeling methods have shown promising results but their performance can suffer from noisy pseudo labels~\cite{liang2020we}. 
Test-time training~\cite{sun19ttt} introduced a self-supervised auxiliary rotation prediction task to be optimized jointly during both source and target training. This approach is limited because it requires altering the source training protocols, which may not be feasible for all models of interest. 
Moreover, the contrastive learning paradigm has been shown to learn more transferable representations compared to rotation prediction as a pre-text task. Recently, \cite{wang2021target} used self-supervised learning in the pre-training stage, however, we argue this method does not fully leverage the strength of self-supervised representation learning during the adaptation stage. 

In this work, we introduce AdaContrast, a novel test-time adaptation strategy that uses
self-supervised contrastive learning on the target domain to exploit the pair-wise information among target samples, which is optimized jointly with pseudo labeling.
Compared to the pretrain-and-finetune paradigm \cite{he2020momentum,chen2020simple,chen2021exploring}, the joint optimization on target domain allows the model to reuse source knowledge to quickly adapt, while benefiting mutually with pseudo labeling. The intuition is that a better target representation facilitates the learning of the decision boundaries \cite{Caron_2018_ECCV}, while the useful priors contained in pseudo labels further enhances the effectiveness of contrastive learning in representation learning. We also show that our auxiliary contrastive learning brings robustness to the pseudo labeling, preventing divergence and allowing the online pseudo labels to consistently provide high-accuracy supervision.

As for the pseudo labeling, we introduce a new online pseudo label refinement scheme that results in generating significantly more correct pseudo labels by using soft k-nearest neighbors voting~\cite{softknn} in the target domain's feature space for each target sample. As shown in Fig. \ref{fig:teaser}, unlike prior works which typically require an offline global memory bank to store pseudo labels/features generated every single or a few epochs~\cite{liang2020we,wang2021target}, we produce and refine pseudo labels at a per-batch basis by aggregating probabilities from nearest neighbors based on feature distances. Relying on a relatively small memory queue instead makes our approach both computationally affordable and suitable for online streaming where target data cannot be revisited such as robotics applications. 

The two key factors in AdaContrast, self-supervised contrastive learning trained jointly with pseudo labeling, offer several empirical merits including hyper parameter insensitivity and better model calibration. Hyper-parameter selection in TTA setting is a key design factor that is often neglected in TTA literature where tuning hyper-parameters is not an option due to lack of access to target labels. We empirically show our proposed AdaContrast approach consistently performs well under a wide range of hyper-parameters. We also found AdaContrast to have a better model calibration \cite{platt1999probabilistic,guo2017calibration} compared to entropy minimization-based methods~\cite{liang2020we}.  We have evaluated our method on major domain adaptation benchmarks where it achieves state-of-the-art test-time adaptation performance. Its 86.8\% average accuracy and 84.5\% overall accuracy on VisDA-C surpass the previous state-of-the-art by +3.8\% and +6.2\%, respectively. We are also the first TTA method to evaluate on the large-scale DomainNet dataset, where AdaContrast achieves state-of-the-art 67.8\% accuracy averaged over 7 domain shifts.

\section{Related Work}
\label{sec:relatedwork}

\noindent\textbf{Domain adaptation} has been extensively explored for many visual tasks, including image classification \cite{tzeng2014deep}, semantic segmentation \cite{tsai2018learning} and object detection \cite{chen2018domain}. The goal of unsupervised domain adaptation (UDA) is to close the performance gap when the source model is deployed on a different target domain without any target annotation. Existing works have made tremendous progress revolving around the idea of feature space alignment, with different mechanisms. \cite{peng2019moment} align the statistics of the distributions, notably the moments at different orders. \cite{long2015learning,tzeng2014deep} exploit maximum mean discrepancy. \cite{ganin2015unsupervised} achieve domain confusion by adversarially training the feature encoder and a domain discriminator, whereas GAN-based methods \cite{hoffman2018cycada} employ generative task to make indistinguishable source and target images. All these methods, however, need to access both source data and target data during the adaptation, making the learning essentially transductive. 

Some recent works on source-free/test-time adaptation focus on the more challenging setting where only source model and unlabeled target data are available \cite{li2020model,kundu2020universal,liang2020we,yang2020unsupervised,li2020model,wang2021tent}.
TENT \cite{wang2021tent} introduce entropy minimization as test-time optimization objective.
SHOT \cite{liang2020we} combined entropy minimization with pseudo labeling. These methods are limited in several aspects. 
First, the entropy minimization objective does not model the relation among different samples and more importantly, 
disrupts the model calibration on target data due to direct entropy optimization. 
Second, the pseudo labels are updated only on a per-epoch basis, 
which fails to reflect the most recent model improvement during an epoch.
In contrast, our method is equipped with contrastive learning for contextual modeling and online pseudo-label for the latest update.

\noindent\textbf{Self-supervised learning} 
methods \cite{noroozi2016unsupervised,larsson2017colorproxy,gidaris2018unsupervised,oord2018representation,Caron_2018_ECCV,chen2020simple,chen2020big,he2020momentum,chen2020improved,chen2021exploring,zbontar2021barlow,grill2020bootstrap} have shown tremendous success in producing transferable visual representations. 
Researchers have found that contrastive-based proxy tasks~\cite{oord2018representation,chen2020simple,chen2020improved,chen2021exploring} can help models to learn a representation that has the potential to replace supervised pre-training (\eg, ImageNet~\cite{deng2009imagenet}).
Consequently, researchers have explored using self-supervised learning for domain adaptation \cite{sun19ttt,sun2019unsupervised,saito2020universal,wang2021target}. ~\cite{sun2019unsupervised,saito2020universal} tackle the unsupervised domain adaptation (UDA) setting where concurrent access to source and target data is allowed.
TTT~\cite{sun19ttt} utilizes rotation prediction task as a proxy to update backbone parameters.
On-target adaptation~\cite{wang2021target} leverages contrastive learning to initialize the target-domain feature, as a separate stage in the proposed framework.
In contrast, we propose a joint learning approach that combines contrastive learning and pseudo labeling.

\noindent\textbf{Pseudo labeling} has been widely adopted in semi-supervised learning \cite{lee2013pseudo,sohn2020fixmatch}, self-supervised learning \cite{Caron_2018_ECCV,asano2019self}, and domain adaptation \cite{liang2020we,liang2021source, liang2021domain, wang2021target}. 
It is a simple yet effective strategy: for unlabeled samples, the predicted labels or cluster assignment are treated as if they were ground truth labels to provide ``supervision''. 
FixMatch~\cite{sohn2020fixmatch} is a semi-supervised learning method benefitting from pseudo labeling and consistency regularization.
The most recent proposed on-target adaptation~\cite{wang2021target} augment the FixMatch-style teacher-student learning with contrastive learned target-domain model.
Our method utilizes weak-strong consistency as a regularizer while additionally denoise pseudo labels via the proposed online refinement \cref{sec:onlinepl}.

\section{Method}
\label{sec:method}
\begin{figure*}[t]
  \centering
   \includegraphics[width=0.8\linewidth]{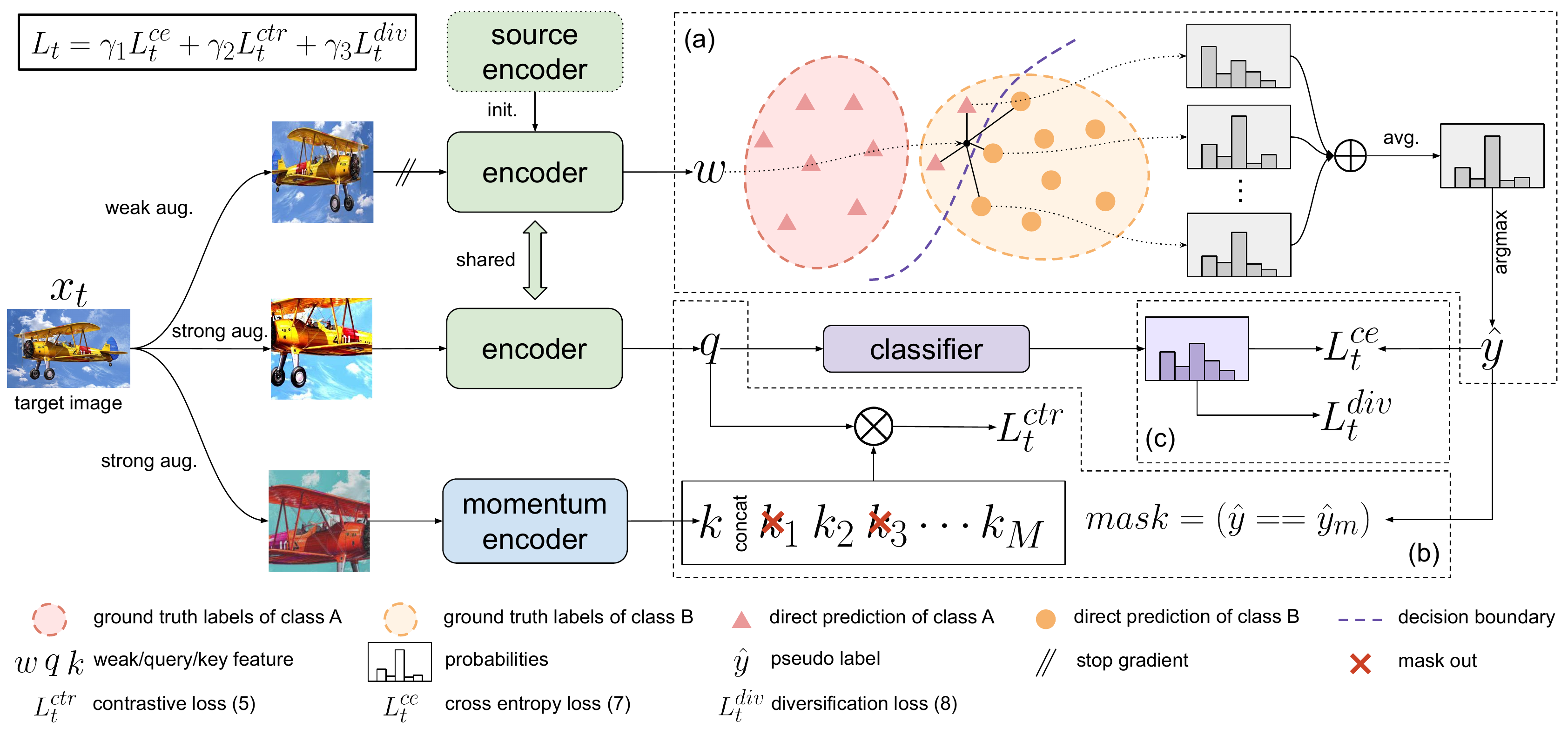}

   \caption{\textbf{Framework of our contrastive test-time adaptation approach (AdaContrast):} In the beginning of adapation, the model and momentum model are initialized by source model. A target image is transformed by one weak and two strong augmentations. (a) The weakly-augmented image is encoded into feature vector $w$ that is used to find nearest neighbors based on cosine distance from the target feature space, which is maintained as a memory queue. The associated probabilities are averaged followed by an argmax to get the refined pseudo label $\hat{y}$ for self-training and contrastive learning. (b) Two strongly augmented versions of the image are encoded into query and key features $q, k$ for momentum contrast \cite{he2020momentum,chen2020improved}, which is applied jointly with self-training. No projection heads are used; current pseudo label and historical pseudo labels are used to exclude same-class negative pairs. (c) The pseudo label $\hat{y}$ obtained from the weakly-augmented image is also used to supervise predictions for the strongly-augmented image, enforcing the weak-strong consistency in the self-training. Diversity regularization is also posed on the same predictions. Note that the queues used for nearest neighbors search and contrastive learning are separate, which are updated (not illustrated here) with $w$ and $k$, respectively.}
   \label{fig:main}
\end{figure*}
We address the closed-set test-time adaptation problem in image classification where source data is not used during the adaptation.  The source model is trained on source pairs of $\{x_s^i, y_s^i\}_{i=1}^{n_s} \in \mathcal{D}_s$ where $x_s^i\in \mathcal{X}_s$ and $y_s^i\in\mathcal{Y}_s$ are images and labels, respectively. Given the trained source model, the goal is to adapt it to unlabeled target data denoted as 
$\{x_t^i\}_{i=1}^{n_t}\in \mathcal{X}_t$. The underlying labels $\{y_t^i\}_{i=1}^{n_t}\in\mathcal{Y}_t$ are accessed only for evaluation purpose. In the closed-set case the source and target domain share the same label space $\mathcal{Y}_s=\mathcal{Y}_t=\mathcal{Y}$. The source model has a general architecture consisting of a feature extractor $f_s(\cdot):\mathcal{X}_s\to\mathbf{R}^D$ and a classifier $h_s(\cdot):\mathbf{R}^D\to\mathbf{R}^C$ where $D$ and $C$ are feature dimension and number of classes. To obtain the source model, we follow \cite{liang2020we} to first train a model on source data with the standard cross-entropy loss $L^{ce}_s = -\sum_{c=1}^{C}\Tilde{y}_s^c\log{p}_s^c$
where $p_s^c=\sigma_c(h_s(f_s((x_s)))$ is the $c$-th element of the model's output after softmax operation $\sigma_c(a)=\frac{\exp(a_c)}{\sum_{k=1}^C\exp{(a_k)}}$, and $\Tilde{y}_s^c$ is the $c$-th element of the converted one-hot label with label-smoothing \cite{labelsmoothing}: $\Tilde{y}_s^c=(1-\alpha)y_s^c+\alpha/C$, where $\alpha=0.1$ is the smoothness coefficient.

In the test-time adaptation phase, we initialize the target model $g_t(\cdot) = h_t(f_t(\cdot))$ with the source model's parameters $\theta_s$.
\subsection{Online pseudo label refinement}
\label{sec:onlinepl}
 During the adaptation, we produce pseudo labels $\{\hat{y}^{i}\}_{i=1}^{n_t}$ for the unlabeled target data using the target model initialized with source weights as a way to re-use knowledge learned from the source domain while gradually bootstrapping to the target domain. 
 Instead of refining and updating pseudo labels only after each epoch \cite{liang2020we,wang2021target}, we propose to predict and refine the pseudo labels at a per-batch basis, so that the model's progressive improvement is reflected in the most recent pseudo labels. The refinement is accomplished via a nearest-neighbor soft voting, which is enabled by a memory queue $Q_w$  representing the target feature space. Specifically, as shown in \cref{fig:main}(a), given a target image $x_t$ and a weak augmentation $t_w$ randomly drawn from a distribution $\mathcal{T}_w$, the weakly-augmented image $t_w(x_t)$ is encoded into a feature vector $w=f_t(t_w(x_t))$, which we use to find its nearest neighbors in the target feature space. The direct prediction for the image is then refined by averaging the probabilities associated with the nearest neighbors, followed by an argmax operation to get the pseudo label $\hat{y}$. Note that this procedure is executed at each mini-batch step.


\paragraph{Memory queue}
To enable the nearest-neighbor search, we maintain a memory queue $Q_w$ of length $M$ storing features and predicted probabilities $\{w'^j, p'^j\}_{j=1}^{M}$  of the weakly-augmented target samples, and update it on-the-fly with the current mini-batch. The memory queue $Q_w$ is initialized with features and probabilities of $M$ randomly selected target samples. Update is done by enqueue and dequeue similar to \cite{he2020momentum}. To make the maintained feature space more stable, we use a slowly changing momentum model $g_t'(\cdot)=h_t'(f_t'(\cdot))$ to calculate update features $w'$ and probabilities $p'$:

\begin{equation}
    w'=f_t'(t_w(x_t)),\quad p'=\sigma{(h_t'(w'))}
\end{equation}
The momentum model $g_t'$'s parameters $\theta_t'$ are initialized with the same source weights $\theta_s$ at the beginning of the adaptation, and updated with momentum $m$ at each mini-batch step instead of back-propagation:

\begin{equation}
    \theta_t'\leftarrow m\theta_t'+(1-m)\theta_t
\end{equation}

\paragraph{Nearest-neighbor soft voting} 
The intuition of soft voting \cite{softknn} is shown in the feature space of \cref{fig:main} (a): the current classifier makes incorrect decisions for some target samples due to domain shift; however, by aggregating knowledge of nearby points we can get a more informed estimate, potentially recovering the correct label. The memory queue $Q_w$ effectively represents our estimate of the evolving target feature space. Therefore, we use the feature vector $w$ of the weakly-augmented image $t_w(x_t)$ to retrieve its $N$ nearest neighbors from $Q_w$ based on the cosine distance between $w$ and the entire set of features $\{w'^j\}_{j=1}^{M}$ stored by $Q_w$. We perform a soft voting among these $N$ neighbors by averaging their probabilities:

\begin{equation}
\label{eq:softvoting}
    \hat{p}^{(i,c)}=\frac{1}{N}\sum_{j\in{\mathcal{I}}_i}p'^{(j,c)}
\end{equation}

where $\mathcal{I}_i$ is the indices of the $N$ nearest neighbors of $w$ in the memory queue $Q_w$. After the voting, we get a less noisy estimate of the categorical probability for target sample, upon which we decide the pseudo label:

\begin{equation}
\label{eq:hardpseudolabel}
    \hat{y}^i=\argmax_c \hat{p}^{(i,c)}
\end{equation}

The obtained pseudo labels will be used in the joint optimization of contrastive learning and self-training, shown in \cref{fig:main} (b) and (c).

\subsection{Joint self-supervised contrastive learning}
\label{sec:contrastivelearning}
Taking inspirations from existing self-supervised contrastive learning works \cite{he2020momentum,chen2020simple,chen2021exploring,grill2020bootstrap} which exploit pair-wise information with contrastive objectives, we apply  self-supervised contrastive learning on target data jointly with self-training during test-time adaptation, as illustrated in \cref{fig:main} (b). In particular, we design our contrastive task following the shared instance-discrimination principle: features of different views of the same image (positive pairs) are pulled closer while features of different images (negative pairs) are pushed away. Different image views are obtained by augmentation: as shown in \cref{fig:main} on the left, given a target image $x_t$, we randomly draw two strong augmentations $t_s, t_s'$ from the same distribution $\mathcal{T}_s$ and augment $x_t$ into two versions $t_s(x_t), t_s'(x_t)$. More specifically, we use MoCo \cite{he2020momentum} as our prototype and introduce several key modifications, which we elaborate next.

\paragraph{Encoder initialization by source} Instead of training the image encoder from scratch for a large number of epochs (usually hundreds) as needed by representation learning \cite{he2020momentum,chen2020simple,grill2020bootstrap}, we reuse the target encoder $f_t$ which is initialized with source model weights. We adopt the momentum encoder $f_t'$ from MoCo as well and initialize it with source weights. We note that this momentum encoder is in fact the same one used for updating memory queue $Q_w$ in \cref{sec:onlinepl}, here reused for producing contrastive features. By reusing knowledge contained in the source weights $\theta_s$, the contrastive learning starts from an informative feature space, therefore requires very few epochs to converge.

\paragraph{Exclusion of same-class negative pairs} The two versions of the target image is encoded into query and key features $q=f_t(t_s(x_t)), k=f_t'(t_s'(x_t))$, respectively. A memory queue $Q_s$ of length $P$ storing features $\{k^j\}_{j=1}^P$ is in turn updated by $k$. The InfoNCE loss applied in MoCo strives to minimize the cosine distance between $q$ and $k$ while maximizing the cosine distances between $q$ and \textit{every} $k^j$ in $Q_s$. Instead, we argue that not pushing away same-class pairs helps learn better semantically meaningful clusters. Specifically, we augment the memory queue $Q_s$ by also storing pseudo labels $\{\hat{y}^j\}_{j=1}^P$ associated with past key features, to exclude same-class pairs from all negative pairs:
\begin{gather}
\label{loss:infonce}
    L_t^{ctr}=L_{\textrm{InfoNCE}}=-\log\frac{\exp{q\cdot k_{+}/\tau}}{\sum_{j\in\mathcal{N}_q}q\cdot k_j/\tau} \\
    \mathcal{N}_q=\{j|1\leq j\leq P, j\in \mathbf{Z}, \hat{y} \neq \hat{y}^j\}\cup\{0\}
\end{gather}

\paragraph{Joint optimization with self-training} While existing self-supervised contrastive learning works \cite{he2020momentum,chen2020simple,grill2020bootstrap} intend to learn transferrable features in a large-scale pre-training stage which is followed by transferring to specific downstream tasks, AdaContrast jointly optimizes the contrastive objective together with self-training in the test-time adaptation phase. Specifically, the modified InfoNCE term \cref{loss:infonce} is combined with the self-training loss in a multi-task fashion (see \cref{loss:totalloss}). The contrastive learning facilitates self-training with better representation, which in turn benefits from the prior brought by more accurate pseudo labels.

\subsection{Additional regularization}
\label{sec:regularization}

\paragraph{Weak-strong consistency} Inspired by FixMatch\cite{sohn2020fixmatch}, we use the pseudo label $\hat{y}$ obtained from the weakly-augmented target image to ``supervise'' the model's prediction for the strongly-augmented version as shown in \cref{fig:main} (c). There are several important distinctions: 1) we do not have access to any ground truth labels, 2) we refine the pseudo labels before using them, 3) we do not apply any confidence thresholding, and 4) our model starts from source initialization. The regularization is reflected in the standard cross entropy loss:

\begin{equation}
    L_t^{ce}=-\mathbb{E}_{x_t\in\mathcal{X}_t}\sum_{c=1}^{C}\hat{y}^c\log{p}_{q}^c
\end{equation}

where ${p}_{q}=\sigma(g_t(t_s(x_t)))$ are the predicted probabilities for the strongly-augmented query image $t_s(x_t)$.

\paragraph{Diversity regularization}
While the online pseudo label refinement introduced in \cref{sec:onlinepl} effectively reduces noises in pseudo labels brought by domain shift, they are still not ideal as the ground truth labels. To prevent the model from blindly trusting the false labels during the adaptation, we use a regularization term in the loss function to encourage class diversification:
\begin{gather}
    L_t^{div}=\mathbb{E}_{x_{t}\in\mathcal{X}_t}\sum_{c=1}^{C}\Bar{p}_{q}^c\log{\Bar{p}}_{q}^c \\
    \Bar{p}_{q}=\mathbb{E}_{x_{t}\in\mathcal{X}_t}\sigma(g_t(t_s(x_{t})))
\end{gather}

This concludes the overall loss function used for training shown in \cref{loss:totalloss}. We set $\gamma_1=\gamma_2=\gamma_3=1.0$ without any tuning for all experiments, showing the merit of hyper-parameter insensitivity:

\begin{equation}
\label{loss:totalloss}
    L_t=\gamma_1L_t^{ce}+\gamma_2L_t^{ctr}+\gamma_3L_t^{div}
\end{equation}


\section{Experiments}
\label{sec:experiments}

We conduct experiments of closed-set adaptation on major benchmarks. In the following, we first compare the proposed AdaContrast with the previous state-of-the-art algorithms. Then, we discuss several desirable test-time properties of AdaContrast, followed by ablation and analysis of the important design elements that brought the gains.

\subsection{Experimental setup}
\label{sec:setup}

\begin{table*}[t!]
\centering
\caption{Classification accuracy (\%) on VisDA-C train $\to$ val. All methods use ResNet-101 backbone except the on-target rows, which use ResNet-18 as student network. Bold is the highest; underline is the second highest. The proposed AdaContrast surpasses the previous state-of-the-art by 3.8\% Avg. When applying an extra knowledge distillation stage following \cite{wang2021target}, we achieve the highest 87.2\% with a small ResNet-18 backbone. AdaContrast also achieves competitive performance of 78.7\% Avg. when used in online adaptation setting.}
\resizebox{1.95\columnwidth}{!}{
\begin{tabular}{cccccccccccccccc}
    \Xhline{1pt}
    Method & source-free  & plane & bcycl & bus & car & horse & knife & mcycl & person & plant & sktbrd & train & truck & \cellcolor{lightgray!30}Avg. \\
    \hline 
    DANN~\cite{ganin2015unsupervised}  & no  & 81.9 & 77.7 & 82.8 & 44.3 & 81.2 & 29.5 & 65.1 & 28.6 & 51.9 & 54.6 & 82.8 & 7.8 & \cellcolor{lightgray!30}57.4 \\
    CDAN~\cite{long2017conditional} & no  & 85.2 & 66.9 & 83.0 & 50.8 & 84.2 & 74.9 & 88.1 & 74.5 & 83.4 & 76.0 & 81.9 & 38.0 & \cellcolor{lightgray!30}73.9\\
    CDAN+BSP~\cite{chen2019transferability} & no  & 92.4 & 61.0 & 81.0 & 57.5 & 89.0 & 80.6 & 90.1 & 77.0 & 84.2 & 77.9 & 82.1 & 38.4 & \cellcolor{lightgray!30}75.9\\
    CAN~\cite{kang2019contrastive} & no  & \underline{97.0} & 87.2 & 82.5 & 74.3 & \textbf{97.8} & \textbf{96.2} & 90.8 & 80.7 & \textbf{96.6} & \textbf{96.3} & 87.5 & \textbf{59.9} & \cellcolor{lightgray!30}\textbf{87.2} \\
    SWD~\cite{lee2019sliced} & no  & 90.8 & 82.5 & 81.7 & 70.5 & 91.7 & 69.5 & 86.3 & 77.5 & 87.4 & 63.6 & 85.6 & 29.2 & \cellcolor{lightgray!30}76.4\\
    MCC~\cite{jin2020minimum} & no & 88.7 & 80.3 & 80.5 & 71.5 & 90.1 & 93.2 & 85.0 & 71.6 & 89.4 & 73.8 & 85.0 & 36.9 & \cellcolor{lightgray!30}78.8 \\
    \hline
    Source only & - & 57.2 & 11.1 & 42.4 & 66.9 & 55.0 & 4.4 & 81.1 & 27.3 & 57.9 & 29.4 & 86.7 & 5.8 & \cellcolor{lightgray!30}43.8 \\
    MA~\cite{li2020model} & yes & 94.8 & 73.4 & 68.8 & 74.8 & 93.1 & \underline{95.4} & 88.6 & 84.7 & 89.1 & 84.7 & 83.5 & 48.1 & \cellcolor{lightgray!30}81.6 \\ 
    BAIT~\cite{yang2020unsupervised} & yes  & 93.7 & 83.2 & 84.5 & 65.0 & 92.9 & \underline{95.4} & 88.1 & 80.8 & 90.0 & 89.0 & 84.0 & 45.3 & \cellcolor{lightgray!30}82.7 \\ 
    SHOT~\cite{liang2020we} & yes  & 95.3 & \underline{87.5} & 78.7 & 55.6 & 94.1 & 94.2 & 81.4 & 80.0 & 91.8 & 90.7 & 86.5 & \underline{59.8} & \cellcolor{lightgray!30}83.0 \\
    + On-target~\cite{wang2021target} & yes & 96.0 & \textbf{89.5} & 84.3 & 67.2 & 95.9 & 94.2 & 91.0 & 81.5 & 93.8 & 89.9 & 89.1 & 58.2 & \cellcolor{lightgray!30}85.9 \\
    AdaContrast (Ours) & yes  & \underline{97.0} & 84.7 & 84.0 & \underline{77.3} & \underline{96.7} & 93.8 & \underline{91.9} & \underline{84.8} & 94.3 & \underline{93.1} & \textbf{94.1} & 49.7 & \cellcolor{lightgray!30}\underline{86.8}  \\
    + On-target~\cite{wang2021target} & yes & \textbf{97.2} & 87.0 & \textbf{86.7} & \textbf{81.7} & 95.5 & 91.6 & \textbf{93.5} & \textbf{86.6} & \underline{95.3} & 90.9 & \underline{92.8} & 47.9 & \cellcolor{lightgray!30}\textbf{87.2} \\
    \hline
    AdaContrast (Ours, online) & yes & 95.0 & 68.0 & 82.7 & 69.6 & 94.3 & 80.8 & 90.3 & 79.6 & 90.6 & 69.7 & 87.6 & 36.0 & \cellcolor{lightgray!30}78.7 \\
    \Xhline{1pt}
\end{tabular}
}
\label{tab:1_visda-c_main}
\end{table*}
\begin{table*}[t!]
\centering
\caption{Classification accuracy (\%) on 7 domain shifts of DomainNet-126. All methods use ResNet-50 backbone. Bold is the highest. The proposed AdaContrast achieves the highest average performance, and on 4 domain shifts. Its performance under online test-time adaptation setting also reaches a competitive number at 62.6\%.}
\resizebox{1.6\columnwidth}{!}{
\begin{tabular}{ccccccccccc}
    \Xhline{1pt}
    Method & Source-free  & R$\to$C & R$\to$P & P$\to$C & C$\to$S & S$\to$P & R$\to$S & P$\to$R & \cellcolor{lightgray!30}Avg. \\
    \hline
    MCC~\cite{jin2020minimum} & no & 44.8  &  65.7   & 41.9 & 34.9 & 47.3 & 35.3 & 72.4 & \cellcolor{lightgray!30}48.9 \\
    \midrule
    Source only & - & 55.5 & 62.7 & 53.0 & 46.9 & 50.1 & 46.3 & 75.0 & \cellcolor{lightgray!30}55.6\\
    TENT~\cite{wang2021tent} & yes & 58.5 & 65.7 & 57.9 & 48.5 & 52.4 & 54.0 & 67.0 & \cellcolor{lightgray!30}57.7 \\
    SHOT \cite{liang2020we} & yes & 67.7 & 68.4 & 66.9 & \textbf{60.1} & \textbf{66.1} & 59.9 & \textbf{80.8} & \cellcolor{lightgray!30}67.1 \\
    AdaContrast (Ours) & yes & \textbf{70.2} & \textbf{69.8} & \textbf{68.6} & 58.0 & 65.9 & \textbf{61.5} & 80.5 & \cellcolor{lightgray!30}\textbf{67.8} \\
    \hline
    AdaContrast (Ours, online) & yes & 61.1 &	66.9 &	60.8 & 53.4	 & 62.7	 & 54.5 &	78.9 &	\cellcolor{lightgray!30}62.6 \\
    \bottomrule
\end{tabular}
}
\label{tab:2_domainnet_main}
\end{table*}

\noindent\textbf{Datasets and Metrics:} We use VisDA-C \cite{peng2017visda} and DomainNet-126 \cite{peng2019moment} for evaluating our method and comparison. Please see the supplemental material for a detailed description for the datasets. It is worth to know that since the original DomainNet has noisy labels, we follow the authors' followup work \cite{saito2019semi} to use a subset of it that contains 126 classes from 4 domains (Real, Sketch, Clipart, Painting), which we refer to as DomainNet-126. We follow \cite{saito2019semi} to evaluate the methods on 7 domain shifts constructed from the 4 domains, and report top-1 accuracy under each domain shift as well as the 7-shift average (denoted Avg.). For VisDA-C we compare the per-class top-1 accuracies, their average (denoted Avg.), and the overall top-1 accuracy (denoted Acc.).

\noindent\textbf{Model Architecture} Our method assumes a general method architecture with a feature encoder followed by a classifier. For comparison purpose, we choose ResNet-18/50/101 models \cite{he2016deep} as our backbones in different experiments. We follow SHOT \cite{liang2020we} to add a 256-dimensional bottleneck consisting of a fully-connected layer followed by a BatchNorm layer \cite{ioffe2015batch} after the backbone, and apply WeightNorm \cite{salimans2016weight} on the classifier. Since a lower dimensional bottleneck is applied, we drop the original projection heads from MoCo \cite{he2020momentum,chen2020improved} without seeing performance drop. 

\noindent\textbf{Baselines} We compare our method with both classical unsupervised domain adaptation (UDA) baselines and source-free/test-time adaptation baselines. For UDA methods we compare to DANN \cite{ganin2015unsupervised}, CDAN \cite{long2017conditional}, CDAN+BSP \cite{chen2019transferability}, CAN \cite{kang2019contrastive}, SWD \cite{lee2019sliced}, and MCC \cite{jin2020minimum}. It is worth noting that all UDA methods have access to source data during adaptation. For TTA methods we compare to MA \cite{li2020model}, BAIT \cite{yang2020unsupervised}, TENT \cite{wang2021tent}, SHOT \cite{liang2020we}, On-target~\cite{wang2021target} as representative methods based on image generation, class prototypes, entropy minimization, pseudo labeling, and the combination of contrastive feature and pseudo labeling. For MCC\footnote{for DomainNet; VisDA-C numbers are cited}, SHOT, and TENT we run the code released by the authors; for other baselines we cite their numbers.

\noindent\textbf{Implementation} We use Pytorch \cite{paszke2019pytorch} for all implementation. \textbf{For source training} we initialize the ResNet backbone with ImageNet-1K \cite{deng2009imagenet} pre-trained weights in the Pytorch model zoo. We follow \cite{liang2020we} to randomly split the source data into 9:1 ratio where 90\% is used to train the source model and 10\% is used for validation. Source training has 10, 60 epochs for VisDA-C and DomainNet-126 respectively. \textbf{For target training} we use only 15 epochs for all datasets unless otherwise noted. \textbf{For all experiments}, we use SGD optimizer with momentum 0.9 and weight decay 1e-4, and cosine annealing on the learning rate, decaying from initial value to zero based on the training progress: $\eta=\eta_0\cdot0.5(\cos{(a\cdot\pi/2)} + 1)$. The newly added bottleneck and classifier layers have learning rate 10 times of the backbone. The initial learning rate for the backbone is set to 2e-4. Batch size is set to 128. 

\subsection{Results}
\label{sec:mainresults}

\textbf{VisDA-C train $\to$ val} \cref{tab:1_visda-c_main} compares AdaContrast with state-of-the-art unsupervised domain adaptation and test-time adaptation methods on VisDA-C's ``train'' to ``val'' shift. For UDA, our method is on-par with a strong UDA baseline CAN \cite{kang2019contrastive} and significantly outperforms the others by a large margin, even though we do not utilize source data at all during test-time adaptation. In the more challenging TTA setting, we achieve the highest per-class average accuracy by a notable margin (+3.8\%) upon SHOT. 
Compared to on-target adaption which is also built with SHOT, we gain an extra 0.9\% improvement, demonstrating the power of joint training and online refinement.
In addition, when applying an extra knowledge distillation phase following \cite{wang2021target}, we are able to reach 87.2\% per-class average with a contrastive (MoCo v2~\cite{chen2020improved}) pre-trained ResNet-18 backbone. 

\textbf{DomainNet-126 seven domain shifts}  \cref{tab:2_domainnet_main} shows the comparison between AdaContrast and state-of-the-art UDA (first section) and TTA (second sections) methods. Without needing source data during the adaptation AdaContrast outperforms the UDA method MCC \cite{jin2020minimum} by +18.9\% on the averaged performance. When being compared to TTA methods AdaContrast outperforms TENT by +10.1\% on the averaged performance. It achieves the best performance on 4 out of 7 domain shifts as well as the highest averaged performance.

\subsection{Analysis and Discussion}

\begin{figure}[t]
	\centering
	\includegraphics[width=0.49\linewidth]{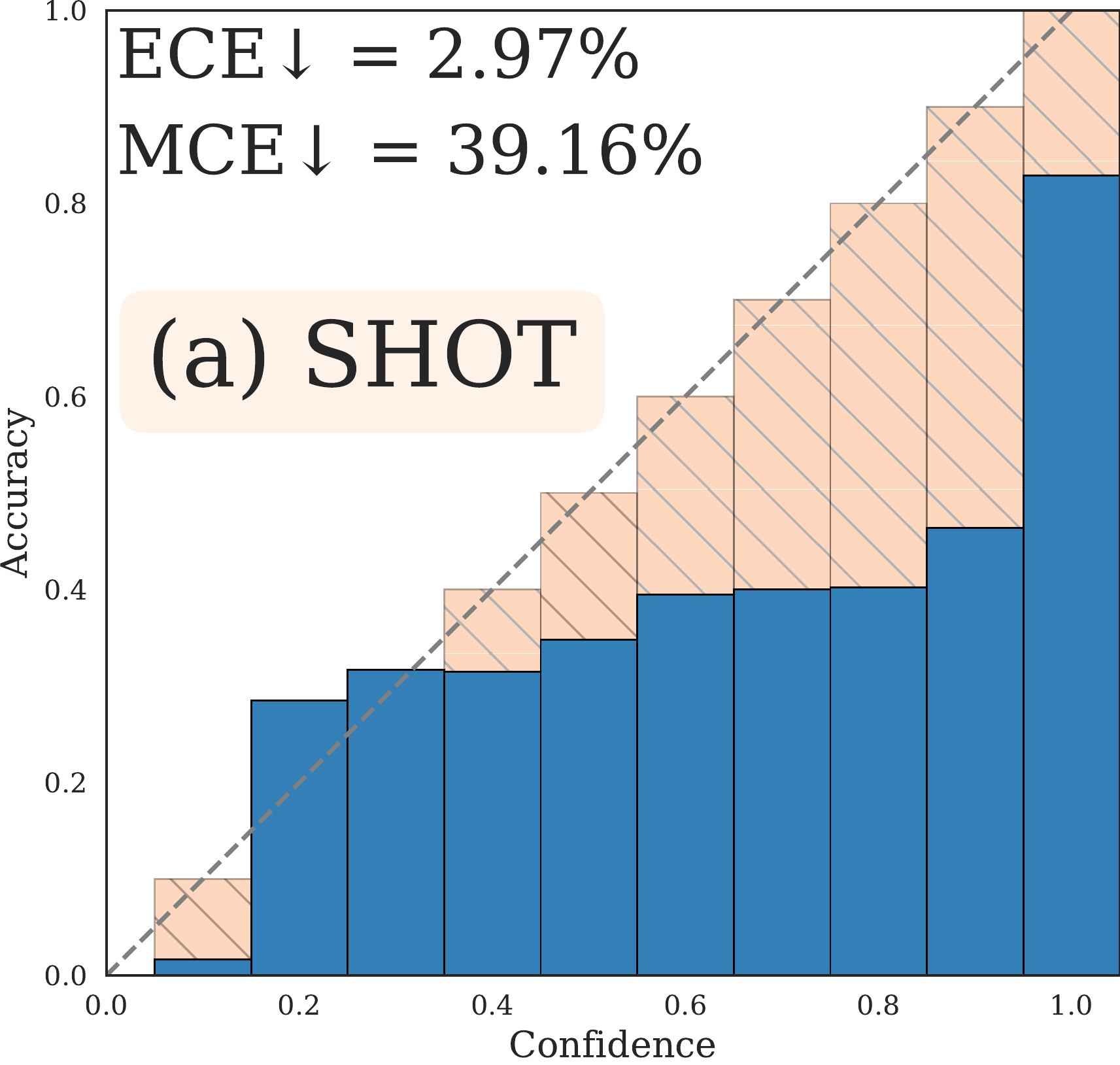}
	\hfill
	\includegraphics[width=0.49\linewidth]{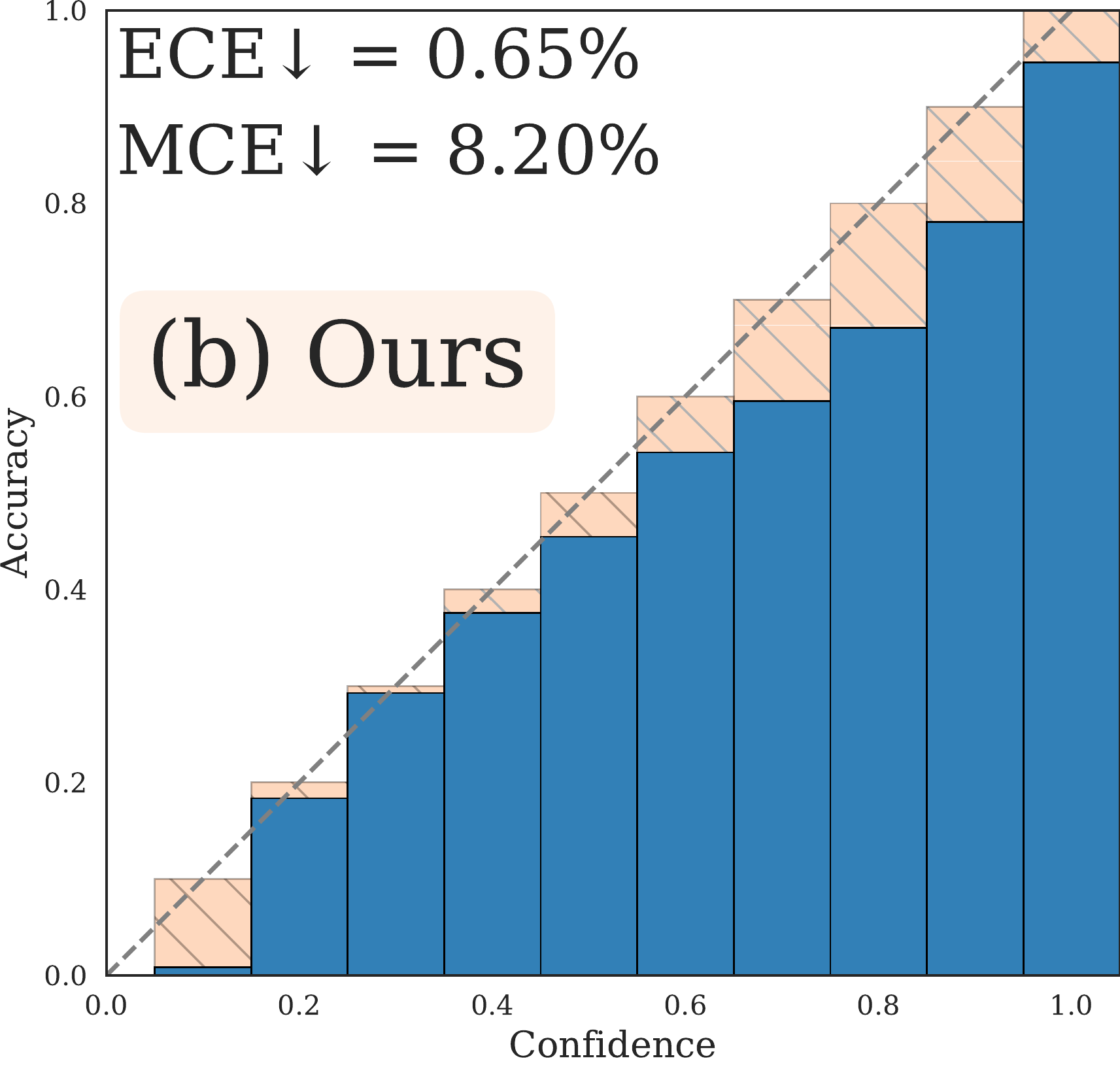}
	\caption{Model calibration comparison of SHOT vs. proposed AdaContrast on VisDA-C validation split after adaptation. The proposed AdaContrast has much better model calibration metrics. For ECE (expected calibration error) and MCE (maximum calibration error): lower is better; the more the bars are aligned with $y=x$ the better.}
	\vspace{-2mm}
	\label{fig:calibration}
\end{figure}
\paragraph{AdaContrast has better model calibration than entropy minimization-based methods.} Entropy minimization-based methods \cite{liang2020we,wang2021tent} achieved competitive results by explicitly making the model ``certain'' on target predictions. However, one setback of this is that the model calibration is disrupted due to direct entropy optimization regardless of true labels. We argue that a good model calibration is an important property of TTA algorithm in practice because it provides a measure to help gauge how much we should trust the adapted model.
In \cref{fig:calibration}, we show the comparison of model calibration for both SHOT and AdaContrast on VisDA-C validation split. We follow the practice of network calibration~\cite{guo2017calibration} and illustrate reliability diagrams~\cite{degroot1983comparison,niculescu2005predicting}.
For each adapted target model, we divide the probability range $[0,1]$ into 10 bins and calculate the model's average accuracy versus average confidence on target data for each bin. The more close the model's bars are to the diagonal line $y=x$, the better calibration it has. As shown in \cref{fig:calibration}, the bars for SHOT significantly falls below $y=x$, meaning its predictions on target data are over-confident, whereas the curve for AdaContrast aligns much better with $y=x$. 
In addition, we calculate two scalar summary statistic of calibration~\cite{naeini2015obtaining}: expected calibration error (ECE) and maximum calibration error (MCE), where the perfectly calibrated classifier should have both scores as low as zero.
Given lower ECE and MCE indicate better calibration, AdaContrast has only 0.65\% ECE and 8.20\% MCE, which decrease by a factor of 4.5+ compared to the over-confident SHOT with 2.97\% ECE and 39.16\% MCE, demonstrating the effectiveness of steering away from entropy minimization.


\begin{figure}[t]
	\centering
	\includegraphics[width=0.49\linewidth]{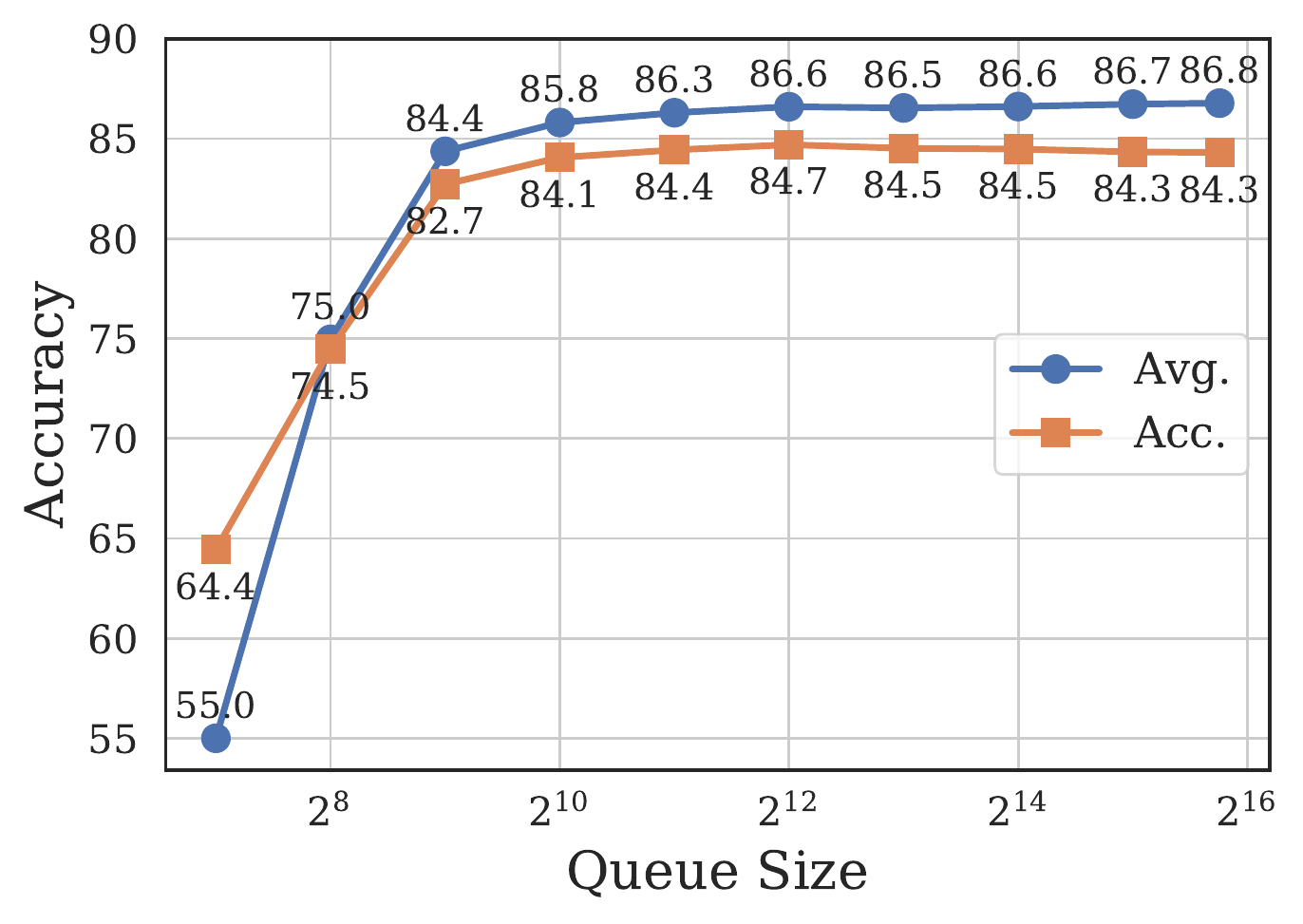}
	\hfill
	\includegraphics[width=0.49\linewidth]{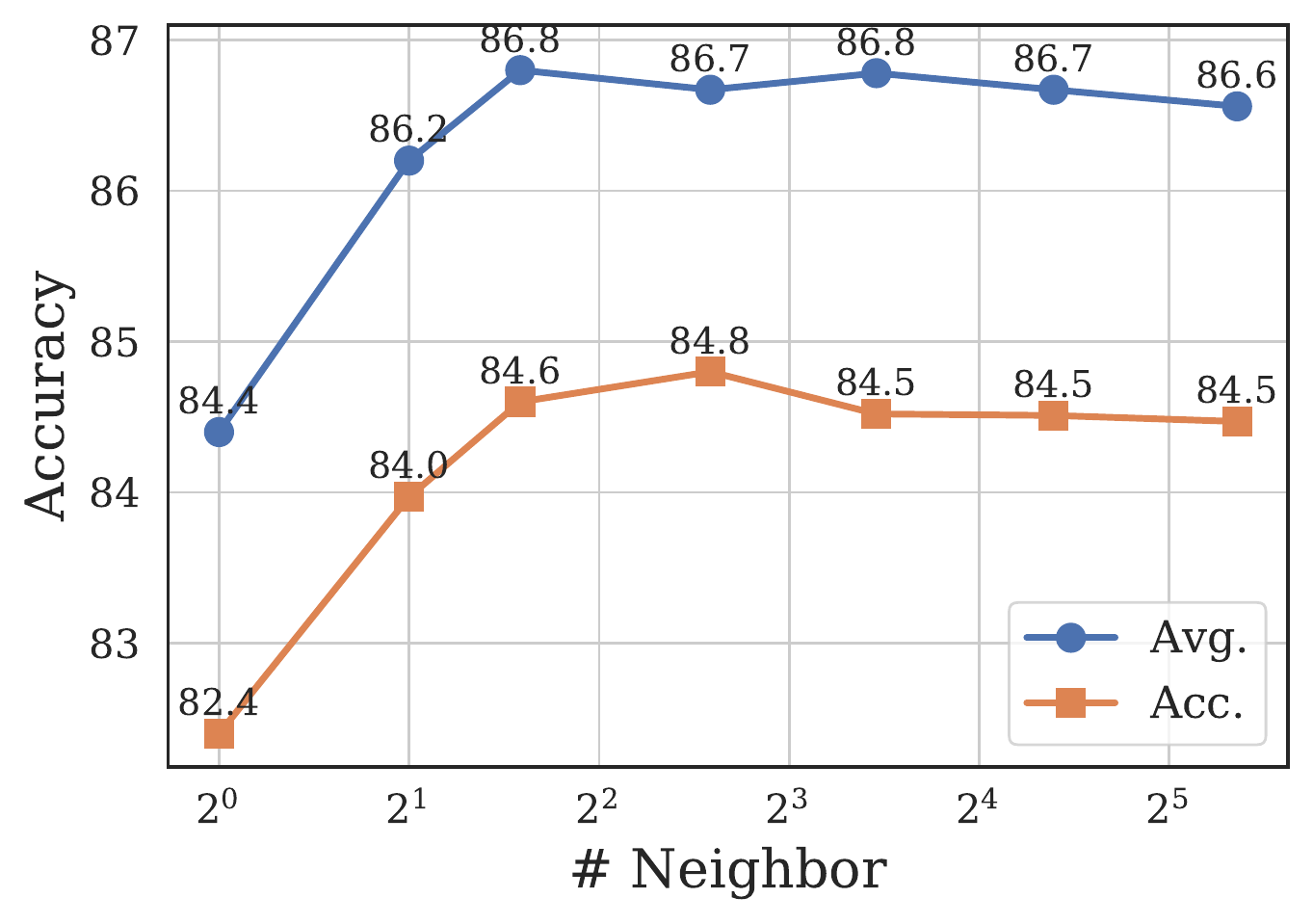}
	\caption{Ablation on memory queue size $M$ and number of neighbors $N$ used in soft voting, on VisDA-C. AdaContrast is able to achieve state-of-the-art performance consistently over a wide range of choices. Notably, $M$ can be as small as less than 4\% of the full dataset and still maintains on-par performance. 
	}
	\vspace{-2mm}
	\label{fig:queue_neighbor}
\end{figure}

\paragraph{AdaContrast is insensitive to hyper-parameters choices.} Hyper-parameter sensitivity is often neglected in TTA literature \cite{liang2020we,kundu2020universal,wang2021tent}, which we believe is an important aspect of TTA algorithms. 
In \cref{fig:queue_neighbor}, we show that under a wide range of hyper-parameters choices that are specific to AdaContrast, the performance is consistently state-of-the-art on VisDA-C. Specifically, we show performance with queue size
$M\in\{128, 256, \cdots, 32768,55388\}$
and number of neighbors in soft voting $N\in\{1,2,3,6,11,21,41\}$.
While we report the strongest results of AdaContrast in \cref{tab:1_visda-c_main} with memory size $M=55388$, queue update, and $N=11$ nearest neighbors for soft voting, as shown in the plots, we see negligible performance degradation when using a much smaller $M$, or varying $N$, achieving around 84.5\% overall accuracy and around 86.7\% per-class averaged accuracy consistently. By using as few as $M=512$ queue size, we are able to reach state-of-the-art TTA performance at 84.4\% per-class average accuray, and on-par (86.3\%) with the full version ($M=55388)$ performance when using $M=2048$, less than 4\% of the full size.
In \cref{tab:5_lr_comparison}, we show AdaContrast is insensitive to learning rate choices as well. With 1x, 3x, 10x the learning rate used for reporting the main results in \cref{sec:mainresults}, AdaContrast consistently achieves state-of-the-art performance on VisDA-C both VisDA-C and DomainNet-126, whereas performance of SHOT \cite{liang2020we} drops noticeably on DomainNet-126 and significantly on VisDA-C.

\noindent\textbf{AdaContrast has strong performance in online test-time adaptation setting.} Since AdaContrast does not rely on global memory banks or processing the entire dataset before the adaptation \cite{liang2020we}, it is naturally suited for online adaptation where target images arrive in a flow of mini-batches and each image is seen only once. Under this setting, we do not decay the learning rate and turn off the pseudo label refinement (note that pseudo labels are still acquired on the fly for each mini-batch, only that the direct predictions are used instead) for the first $X$ samples, and turn it on once the memory queue $Q_w$ has accumulated $X$ feature-probability pairs. We emprically show that with $X=2048$ for VisDA-C (less than 4\% of the entire dataset) and $X=1024$ (less than 4\% on the entire datasets on average), we are able to achieve state-of-the-art online adaptation performance by large margins. 
AdaContrast achieves 62.6\% accuracy averaged on 7 domain shifts in \textbf{DomainNet-126}, surpassing the UDA method MCC \cite{jin2020minimum} (see \cref{tab:2_domainnet_main}) by +13.7\%. On \textbf{VisDA-C}, AdaContrast's impressive 78.7\% overall accuracy slightly surpasses the performance of the offline SHOT \cite{liang2020we} by +0.5\%, the 78.7\% per-class average accuracy surpassing 4 UDA methods listed in \cref{tab:1_visda-c_main}.

\begin{table}[t!]
\centering
\caption{Comparison of classification accuracy (\%) on DomainNet-126 and VisDA-C between AdaContrast and SHOT under 1x, 3x, and 10x learning rate scaling. AdaContrast is less sensitive to the choice of learning rate, achieving consistently high performance on both datasets.}
\resizebox{0.85\columnwidth}{!}{
\begin{tabular}{c|c|c|cc}
    \Xhline{1pt}
    \multicolumn{1}{c|}{\multirow{2}{*}{Method}} & \multicolumn{1}{c|}{\multirow{2}{*}{lr scale}} & DN-126 & \multicolumn{2}{c}{VisDA-C} \\
\multicolumn{1}{c|}{} &  & Avg. & Acc. & Avg. \\ \hline
 \multirow{3}{*}{\begin{tabular}[c]{@{}c@{}}SHOT\\ \cite{liang2020we}\end{tabular}} & $1\times$ & 67.1 & 78.3 & 83.0 \\
  & $3\times$ & 66.4  & 77.6 & 82.2  \\
 & $10\times$ & 64.7 & 66.8 &  72.1 \\ \hline
 \multirow{3}{*}{\begin{tabular}[c]{@{}c@{}}AdaContrast\\ (Ours)\end{tabular}} & $1\times$ & 67.8 & 84.5 & 86.8 \\
 & $3\times$ & 67.8 & 84.7 & 86.8 \\
  & $10\times$ & 67.5 & 85.0 & 86.6  \\ 
    \bottomrule
\end{tabular}
}
\label{tab:5_lr_comparison}
\end{table}

\subsection{Ablation studies}
\label{sec:ablationstudies}
\begin{table}[t!]
\centering
\caption{Ablation study of algorithmic components of proposed AdaContrast measured by classification accuracy (\%) on DomainNet-126 under 1x learning rate scaling and VisDA-C under both 1x, 10x learning rate scaling. \#0 for source-only baseline to start with. Online pseudo label refinement \cref{sec:onlinepl}, joint contrastive learning \cref{sec:contrastivelearning} and regularzation techniques \cref{sec:regularization} are able to bring significant performance gain as well as hyper-parameter insensitivity.}
\resizebox{1.0\columnwidth}{!}{
\begin{tabular}{cccccccc}
\Xhline{1pt}
\multicolumn{1}{c}{\multirow{2}{*}{\#}} & \multirow{2}{*}{\begin{tabular}[c]{@{}l@{}}Pseudo\\ labeling\end{tabular}}  & \multirow{2}{*}{\begin{tabular}[c]{@{}l@{}}Online\\ pl. ref\end{tabular}}  & \multirow{2}{*}{\begin{tabular}[c]{@{}l@{}}Joint\\ctr.\end{tabular}} & \multicolumn{1}{c}{\multirow{2}{*}{Reg.}} & \multicolumn{1}{c}{DN-126} & \multicolumn{1}{c}{VisDA-C} & \multicolumn{1}{c}{VisDA-C}                        \\ 
 & \multicolumn{1}{c}{}   & \multicolumn{1}{c}{}   & \multicolumn{1}{c}{}   & \multicolumn{1}{c}{}     & \multicolumn{1}{c}{(lr1x)}  & \multicolumn{1}{c}{(lr1x)} &          \multicolumn{1}{c}{(lr10x)} \\
 \midrule
 0 &                &            &            &            & 55.6   & 43.8 & 43.8 \\
 1 & \checkmark     &            &            &            & 58.5   & 55.0 & 44.5 \\
 2 & \checkmark     & \checkmark &            &            & 64.7   & 86.5 & 9.9  \\ 
 3 & \checkmark     & \checkmark & \checkmark &            & 67.9   & 85.7 & 84.3 \\ 
 4 & \checkmark     & \checkmark & \checkmark & \checkmark & 67.8   & 86.8 & 86.6 \\
 \bottomrule
\end{tabular}
}
\label{tab:3_ablations}
\end{table}

In \cref{tab:3_ablations}, we start with applying the simplest form of pseudo labeling (referred as \#1), which makes inference on the entire target dataset at the beginning of each epoch and takes all predictions as pseudo labels for the epoch. This achieves 58.5\% average accuracy (Avg.) on DomainNet-126, 55.0\% per-class averaged accuracy (Avg.) on VisDA-C with 1x learning rate, but merely 44.5\% Avg. with 10x learning rate. We note that for VisDA-C we include expriments with 10x learning rate of that used in reporting the main results in \cref{tab:1_visda-c_main}, to emphasize the effect of each component of AdaContrast under an unfortunate choice of learning rate, diving deeper into observations from \cref{tab:5_lr_comparison}.

\textbf{Online pseudo label refinement} In row \#2 we change the pseudo labeling scheme to the one introduced in \ref{sec:onlinepl}. Due to having more accurate pseudo labels, the performance on DomainNet-126 increases by +6.2\% to 64.7\% and significantly by +31.5\% to 86.5\% on VisDA-C with 1x learning rate. However, switching to the online refinement scheme is not trivial, since it is prone to diverge due to bad hyper-parameter selection and compounding errors. As shown in the VisDA-C 10x learning rate performance where the accuracies drop significantly down to near random (12 classes). However, the cross-entropy loss did not diverge from our observation, which means the model severely overfitted to the highly-noisy pseudo labels which we are unable to know.

\textbf{Joint self-supervised contrastive learning} In row \#3 we show results obtained by enabling the joint contrastive learning introduced in \ref{sec:contrastivelearning}, which is simply done by setting $\gamma_2=1.0$ for $L_t^{ctr}$ in \cref{loss:totalloss}. This brings another significant performance gain on DomainNet-126, from 64.7\% average accuracy to 67.9\%. Notably on VisDA-C with 10x learning rate, the joint contrastive learning is able to recover the diverged accuracy from 10.0\% in row \#2 to 84.3\% Avg. This demonstrates the huge potential of our joint contrastive learning in stablizing the feature space, therefore ensuring the model is less susceptible to the compounding errors in pseudo labels as well as hyper-parameter choices. It is worth noting that the improvements include gains from using pseudo labels to exclude same-class negatives (\cref{sec:contrastivelearning}), on top of 67.7\% (+0.2\%), 83.6\% (+2.1\%), and 81.5 (+2.8\%) for the three entries of DomainNet-126 and VisDA-C without excluding same-class negatives. This validates the effectiveness of using semantic priors in pseudo labels to benefit contrastive learning.

\textbf{Diversity and weak-strong regularization} In row \#4 and  we show the effect of two additional regularization in \cref{sec:regularization}: the weak-strong consistency and diversity term $L_t^{div}$. On DomainNet-126 they keep the model's high performance around 67.8\% consistently, whereas on VisDA-C with 10x learning rate they bring further improvements: we get +0.9\%  gains on per-class average accuracy.


\section{Limitations}
Domain adaptation methods are foundational, and as such have as much potential for misuse as they have for beneficial application. Adaptation methods have the potential to increase the robustness of models deployed to new domains, which could amplify the benefits and harms of larger AI applications. Our method improves model calibration, which in general provides for more reliable systems; however this could lead to inappropriate trust in deployed systems.

\section{Conclusion}
We introduced AdaContrast, a novel test-time adaptation approach for closed-set DA in image classification. AdaContrast starts from a pretrained model on the source domain and uses contrastive learning along with pseudo labeling on the target domain. We proposed an online refinement scheme that generates pseudo labels in a per-batch basis and refines the predictions using nearest neighbor soft voting technique which results in significantly more accurate pseudo labels. We showed AdaContrast not only surpassed the existing TTA approaches on major DA benchmarks but also has several empirical merits: hyper-parameter insensitivity, better model calibration, and no need for global memory banks, which we believe are all desirable properties of successful TTA algorithms.

{\small
\bibliographystyle{ieee_fullname}
\bibliography{egbib}

\begin{thebibliography}{10}\itemsep=-1pt

\bibitem{asano2019self}
Yuki~Markus Asano, Christian Rupprecht, and Andrea Vedaldi.
\newblock Self-labelling via simultaneous clustering and representation
  learning.
\newblock {\em arXiv preprint arXiv:1911.05371}, 2019.

\bibitem{bousmalis2017unsupervised}
Konstantinos Bousmalis, Nathan Silberman, David Dohan, Dumitru Erhan, and Dilip
  Krishnan.
\newblock Unsupervised pixel-level domain adaptation with generative
  adversarial networks.
\newblock In {\em Proceedings of the IEEE conference on computer vision and
  pattern recognition}, pages 3722--3731, 2017.

\bibitem{Caron_2018_ECCV}
Mathilde Caron, Piotr Bojanowski, Armand Joulin, and Matthijs Douze.
\newblock Deep clustering for unsupervised learning of visual features.
\newblock In {\em Proceedings of the European Conference on Computer Vision
  (ECCV)}, September 2018.

\bibitem{chen2020simple}
Ting Chen, Simon Kornblith, Mohammad Norouzi, and Geoffrey Hinton.
\newblock A simple framework for contrastive learning of visual
  representations.
\newblock In {\em International conference on machine learning}, pages
  1597--1607. PMLR, 2020.

\bibitem{chen2020big}
Ting Chen, Simon Kornblith, Kevin Swersky, Mohammad Norouzi, and Geoffrey
  Hinton.
\newblock Big self-supervised models are strong semi-supervised learners.
\newblock {\em arXiv preprint arXiv:2006.10029}, 2020.

\bibitem{chen2020improved}
Xinlei Chen, Haoqi Fan, Ross Girshick, and Kaiming He.
\newblock Improved baselines with momentum contrastive learning.
\newblock {\em arXiv preprint arXiv:2003.04297}, 2020.

\bibitem{chen2021exploring}
Xinlei Chen and Kaiming He.
\newblock Exploring simple siamese representation learning.
\newblock In {\em Proceedings of the IEEE/CVF Conference on Computer Vision and
  Pattern Recognition}, pages 15750--15758, 2021.

\bibitem{chen2019transferability}
Xinyang Chen, Sinan Wang, Mingsheng Long, and Jianmin Wang.
\newblock Transferability vs. discriminability: Batch spectral penalization for
  adversarial domain adaptation.
\newblock In {\em International conference on machine learning}, pages
  1081--1090. PMLR, 2019.

\bibitem{chen2018domain}
Yuhua Chen, Wen Li, Christos Sakaridis, Dengxin Dai, and Luc Van~Gool.
\newblock Domain adaptive faster r-cnn for object detection in the wild.
\newblock In {\em Proceedings of the IEEE conference on computer vision and
  pattern recognition}, pages 3339--3348, 2018.

\bibitem{degroot1983comparison}
Morris~H DeGroot and Stephen~E Fienberg.
\newblock The comparison and evaluation of forecasters.
\newblock {\em Journal of the Royal Statistical Society: Series D (The
  Statistician)}, 32(1-2):12--22, 1983.

\bibitem{deng2009imagenet}
Jia Deng, Wei Dong, Richard Socher, Li-Jia Li, Kai Li, and Li Fei-Fei.
\newblock Imagenet: A large-scale hierarchical image database.
\newblock In {\em 2009 IEEE conference on computer vision and pattern
  recognition}, pages 248--255. Ieee, 2009.

\bibitem{ganin2015unsupervised}
Yaroslav Ganin and Victor Lempitsky.
\newblock Unsupervised domain adaptation by backpropagation.
\newblock In {\em International conference on machine learning}, pages
  1180--1189. PMLR, 2015.

\bibitem{ganin2016domain}
Yaroslav Ganin, Evgeniya Ustinova, Hana Ajakan, Pascal Germain, Hugo
  Larochelle, Fran{\c{c}}ois Laviolette, Mario Marchand, and Victor Lempitsky.
\newblock Domain-adversarial training of neural networks.
\newblock {\em The journal of machine learning research}, 17(1):2096--2030,
  2016.

\bibitem{gidaris2018unsupervised}
Spyros Gidaris, Praveer Singh, and Nikos Komodakis.
\newblock Unsupervised representation learning by predicting image rotations.
\newblock {\em arXiv preprint arXiv:1803.07728}, 2018.

\bibitem{grill2020bootstrap}
Jean-Bastien Grill, Florian Strub, Florent Altch{\'e}, Corentin Tallec,
  Pierre~H Richemond, Elena Buchatskaya, Carl Doersch, Bernardo~Avila Pires,
  Zhaohan~Daniel Guo, Mohammad~Gheshlaghi Azar, et~al.
\newblock Bootstrap your own latent: A new approach to self-supervised
  learning.
\newblock {\em arXiv preprint arXiv:2006.07733}, 2020.

\bibitem{guo2017calibration}
Chuan Guo, Geoff Pleiss, Yu Sun, and Kilian~Q Weinberger.
\newblock On calibration of modern neural networks.
\newblock In {\em International Conference on Machine Learning}, pages
  1321--1330. PMLR, 2017.

\bibitem{he2020momentum}
Kaiming He, Haoqi Fan, Yuxin Wu, Saining Xie, and Ross Girshick.
\newblock Momentum contrast for unsupervised visual representation learning.
\newblock In {\em Proceedings of the IEEE/CVF Conference on Computer Vision and
  Pattern Recognition}, pages 9729--9738, 2020.

\bibitem{he2016deep}
Kaiming He, Xiangyu Zhang, Shaoqing Ren, and Jian Sun.
\newblock Deep residual learning for image recognition.
\newblock In {\em Proceedings of the IEEE conference on computer vision and
  pattern recognition}, pages 770--778, 2016.

\bibitem{hoffman2018cycada}
Judy Hoffman, Eric Tzeng, Taesung Park, Jun-Yan Zhu, Phillip Isola, Kate
  Saenko, Alexei Efros, and Trevor Darrell.
\newblock Cycada: Cycle-consistent adversarial domain adaptation.
\newblock In {\em International conference on machine learning}, pages
  1989--1998. PMLR, 2018.

\bibitem{ioffe2015batch}
Sergey Ioffe and Christian Szegedy.
\newblock Batch normalization: Accelerating deep network training by reducing
  internal covariate shift.
\newblock In {\em International conference on machine learning}, pages
  448--456. PMLR, 2015.

\bibitem{jin2020minimum}
Ying Jin, Ximei Wang, Mingsheng Long, and Jianmin Wang.
\newblock Minimum class confusion for versatile domain adaptation.
\newblock In {\em European Conference on Computer Vision}, pages 464--480.
  Springer, 2020.

\bibitem{kang2019contrastive}
Guoliang Kang, Lu Jiang, Yi Yang, and Alexander~G Hauptmann.
\newblock Contrastive adaptation network for unsupervised domain adaptation.
\newblock In {\em Proceedings of the IEEE/CVF Conference on Computer Vision and
  Pattern Recognition}, pages 4893--4902, 2019.

\bibitem{kundu2020universal}
Jogendra~Nath Kundu, Naveen Venkat, R~Venkatesh Babu, et~al.
\newblock Universal source-free domain adaptation.
\newblock In {\em Proceedings of the IEEE/CVF Conference on Computer Vision and
  Pattern Recognition}, pages 4544--4553, 2020.

\bibitem{larsson2017colorproxy}
Gustav Larsson, Michael Maire, and Gregory Shakhnarovich.
\newblock Colorization as a proxy task for visual understanding.
\newblock In {\em CVPR}, 2017.

\bibitem{lee2019sliced}
Chen-Yu Lee, Tanmay Batra, Mohammad~Haris Baig, and Daniel Ulbricht.
\newblock Sliced wasserstein discrepancy for unsupervised domain adaptation.
\newblock In {\em Proceedings of the IEEE/CVF Conference on Computer Vision and
  Pattern Recognition}, pages 10285--10295, 2019.

\bibitem{lee2013pseudo}
Dong-Hyun Lee et~al.
\newblock Pseudo-label: The simple and efficient semi-supervised learning
  method for deep neural networks.
\newblock In {\em Workshop on challenges in representation learning, ICML},
  volume~3, page 896, 2013.

\bibitem{li2020model}
Rui Li, Qianfen Jiao, Wenming Cao, Hau-San Wong, and Si Wu.
\newblock Model adaptation: Unsupervised domain adaptation without source data.
\newblock In {\em Proceedings of the IEEE/CVF Conference on Computer Vision and
  Pattern Recognition}, pages 9641--9650, 2020.

\bibitem{liang2020we}
Jian Liang, Dapeng Hu, and Jiashi Feng.
\newblock Do we really need to access the source data? source hypothesis
  transfer for unsupervised domain adaptation.
\newblock In {\em International Conference on Machine Learning}, pages
  6028--6039. PMLR, 2020.

\bibitem{liang2021domain}
Jian Liang, Dapeng Hu, and Jiashi Feng.
\newblock Domain adaptation with auxiliary target domain-oriented classifier.
\newblock In {\em Proceedings of the IEEE/CVF Conference on Computer Vision and
  Pattern Recognition}, pages 16632--16642, 2021.

\bibitem{liang2021source}
Jian Liang, Dapeng Hu, Yunbo Wang, Ran He, and Jiashi Feng.
\newblock Source data-absent unsupervised domain adaptation through hypothesis
  transfer and labeling transfer.
\newblock {\em IEEE Transactions on Pattern Analysis and Machine Intelligence},
  2021.

\bibitem{long2015learning}
Mingsheng Long, Yue Cao, Jianmin Wang, and Michael Jordan.
\newblock Learning transferable features with deep adaptation networks.
\newblock In {\em International conference on machine learning}, pages 97--105.
  PMLR, 2015.

\bibitem{long2017conditional}
Mingsheng Long, Zhangjie Cao, Jianmin Wang, and Michael~I Jordan.
\newblock Conditional adversarial domain adaptation.
\newblock {\em arXiv preprint arXiv:1705.10667}, 2017.

\bibitem{softknn}
HB Mitchell and PA Schaefer.
\newblock A “soft” k-nearest neighbor voting scheme.
\newblock {\em International journal of intelligent systems}, 16(4):459--468,
  2001.

\bibitem{naeini2015obtaining}
Mahdi~Pakdaman Naeini, Gregory Cooper, and Milos Hauskrecht.
\newblock Obtaining well calibrated probabilities using bayesian binning.
\newblock In {\em Twenty-Ninth AAAI Conference on Artificial Intelligence},
  2015.

\bibitem{niculescu2005predicting}
Alexandru Niculescu-Mizil and Rich Caruana.
\newblock Predicting good probabilities with supervised learning.
\newblock In {\em Proceedings of the 22nd international conference on Machine
  learning}, pages 625--632, 2005.

\bibitem{noroozi2016unsupervised}
Mehdi Noroozi and Paolo Favaro.
\newblock Unsupervised learning of visual representations by solving jigsaw
  puzzles.
\newblock In {\em European conference on computer vision}, pages 69--84.
  Springer, 2016.

\bibitem{oord2018representation}
Aaron van~den Oord, Yazhe Li, and Oriol Vinyals.
\newblock Representation learning with contrastive predictive coding.
\newblock {\em arXiv preprint arXiv:1807.03748}, 2018.

\bibitem{paszke2019pytorch}
Adam Paszke, Sam Gross, Francisco Massa, Adam Lerer, James Bradbury, Gregory
  Chanan, Trevor Killeen, Zeming Lin, Natalia Gimelshein, Luca Antiga, et~al.
\newblock Pytorch: An imperative style, high-performance deep learning library.
\newblock {\em Advances in neural information processing systems},
  32:8026--8037, 2019.

\bibitem{peng2019moment}
Xingchao Peng, Qinxun Bai, Xide Xia, Zijun Huang, Kate Saenko, and Bo Wang.
\newblock Moment matching for multi-source domain adaptation.
\newblock In {\em Proceedings of the IEEE International Conference on Computer
  Vision}, pages 1406--1415, 2019.

\bibitem{peng2017visda}
Xingchao Peng, Ben Usman, Neela Kaushik, Judy Hoffman, Dequan Wang, and Kate
  Saenko.
\newblock Visda: The visual domain adaptation challenge.
\newblock {\em arXiv preprint arXiv:1710.06924}, 2017.

\bibitem{platt1999probabilistic}
John Platt et~al.
\newblock Probabilistic outputs for support vector machines and comparisons to
  regularized likelihood methods.
\newblock {\em Advances in large margin classifiers}, 10(3):61--74, 1999.

\bibitem{datasetshift}
Joaquin Qui{\~n}onero-Candela, Masashi Sugiyama, Neil~D Lawrence, and Anton
  Schwaighofer.
\newblock {\em Dataset shift in machine learning}.
\newblock MIT Press, 2009.

\bibitem{covariateshift}
Joaquin Qui{\~n}onero-Candela, Masashi Sugiyama, Anton Schwaighofer, and N
  Lawrence.
\newblock Covariate shift and local learning by distribution matching, 2008.

\bibitem{saenko2010adapting}
Kate Saenko, Brian Kulis, Mario Fritz, and Trevor Darrell.
\newblock Adapting visual category models to new domains.
\newblock In {\em European conference on computer vision}, pages 213--226.
  Springer, 2010.

\bibitem{saito2019semi}
Kuniaki Saito, Donghyun Kim, Stan Sclaroff, Trevor Darrell, and Kate Saenko.
\newblock Semi-supervised domain adaptation via minimax entropy.
\newblock {\em ICCV}, 2019.

\bibitem{saito2020universal}
Kuniaki Saito, Donghyun Kim, Stan Sclaroff, and Kate Saenko.
\newblock Universal domain adaptation through self supervision.
\newblock {\em arXiv preprint arXiv:2002.07953}, 2020.

\bibitem{salimans2016weight}
Tim Salimans and Durk~P Kingma.
\newblock Weight normalization: A simple reparameterization to accelerate
  training of deep neural networks.
\newblock {\em Advances in neural information processing systems}, 29:901--909,
  2016.

\bibitem{sohn2020fixmatch}
Kihyuk Sohn, David Berthelot, Chun-Liang Li, Zizhao Zhang, Nicholas Carlini,
  Ekin~D Cubuk, Alex Kurakin, Han Zhang, and Colin Raffel.
\newblock Fixmatch: Simplifying semi-supervised learning with consistency and
  confidence.
\newblock {\em arXiv preprint arXiv:2001.07685}, 2020.

\bibitem{sun2019unsupervised}
Yu Sun, Eric Tzeng, Trevor Darrell, and Alexei~A Efros.
\newblock Unsupervised domain adaptation through self-supervision.
\newblock {\em arXiv preprint arXiv:1909.11825}, 2019.

\bibitem{sun19ttt}
Yu Sun, Xiaolong Wang, Liu Zhuang, John Miller, Moritz Hardt, and Alexei~A.
  Efros.
\newblock Test-time training with self-supervision for generalization under
  distribution shifts.
\newblock In {\em ICML}, 2020.

\bibitem{labelsmoothing}
Christian Szegedy, Vincent Vanhoucke, Sergey Ioffe, Jon Shlens, and Zbigniew
  Wojna.
\newblock Rethinking the inception architecture for computer vision.
\newblock In {\em Proceedings of the IEEE conference on computer vision and
  pattern recognition}, pages 2818--2826, 2016.

\bibitem{tsai2018learning}
Yi-Hsuan Tsai, Wei-Chih Hung, Samuel Schulter, Kihyuk Sohn, Ming-Hsuan Yang,
  and Manmohan Chandraker.
\newblock Learning to adapt structured output space for semantic segmentation.
\newblock In {\em Proceedings of the IEEE conference on computer vision and
  pattern recognition}, pages 7472--7481, 2018.

\bibitem{tzeng2017adversarial}
Eric Tzeng, Judy Hoffman, Kate Saenko, and Trevor Darrell.
\newblock Adversarial discriminative domain adaptation.
\newblock In {\em Proceedings of the IEEE conference on computer vision and
  pattern recognition}, pages 7167--7176, 2017.

\bibitem{tzeng2014deep}
Eric Tzeng, Judy Hoffman, Ning Zhang, Kate Saenko, and Trevor Darrell.
\newblock Deep domain confusion: Maximizing for domain invariance.
\newblock {\em arXiv preprint arXiv:1412.3474}, 2014.

\bibitem{wang2021target}
Dequan Wang, Shaoteng Liu, Sayna Ebrahimi, Evan Shelhamer, and Trevor Darrell.
\newblock On-target adaptation.
\newblock {\em arXiv preprint arXiv:2109.01087}, 2021.

\bibitem{wang2021tent}
Dequan Wang, Evan Shelhamer, Shaoteng Liu, Bruno Olshausen, and Trevor Darrell.
\newblock Tent: Fully test-time adaptation by entropy minimization.
\newblock In {\em International Conference on Learning Representations}, 2021.

\bibitem{yang2020unsupervised}
Shiqi Yang, Yaxing Wang, Joost van~de Weijer, Luis Herranz, and Shangling Jui.
\newblock Unsupervised domain adaptation without source data by casting a bait.
\newblock {\em arXiv preprint arXiv:2010.12427}, 2020.

\bibitem{yang2021generalized}
Shiqi Yang, Yaxing Wang, Joost van~de Weijer, Luis Herranz, and Shangling Jui.
\newblock Generalized source-free domain adaptation.
\newblock In {\em Proceedings of the IEEE/CVF International Conference on
  Computer Vision}, pages 8978--8987, 2021.

\bibitem{zbontar2021barlow}
Jure Zbontar, Li Jing, Ishan Misra, Yann LeCun, and St{\'e}phane Deny.
\newblock Barlow twins: Self-supervised learning via redundancy reduction.
\newblock {\em arXiv preprint arXiv:2103.03230}, 2021.

\end{thebibliography}
}

\end{document}


\title{
Contrastive Test-time Adaptation \\
-- Supplementary Material --
}

\author{First Author\\
Institution1\\
Institution1 address\\
{\tt\small firstauthor@i1.org}
\and
Second Author\\
Institution2\\
First line of institution2 address\\
{\tt\small secondauthor@i2.org}
}
\maketitle


\section{Algorithm Pseudo Code}
We summarize the AdaContrast algorithm in pseudo code in \cref{alg:adacontrast}.

\begin{algorithm}
\caption{AdaContrast Pseudocode}\label{alg:adacontrast}
\begin{algorithmic}
\Require target data $\mathcal{X}_t$, source weights $\theta_s$, target model $g_t(\cdot)$

\State Initialize $g_t(\cdot)$ with $\theta_s$.
\State Create momentum model $g_t'(\cdot)$ and initialize it with $\theta_s$.
\State Initialize $Q_w=\{\}.$ Randomly initialize $Q_s=\{k^j\}_{j=1}^P$.
\State Randomly sample $M$ images $\mathcal{X}_t'=\{x_t^i\}_{i\in\mathcal{I}_M},|\mathcal{I}_M|=M$. \Comment{Initialize memory queue $Q_w$}
\For {$x_t^i$ in $\mathcal{X}_t'$} 
    \State Draw $t_w\sim\mathcal{T}_w$. Transform $x_w=t_w(x_t^i)$.
    \State $w'=f_t'(x_w),p'=\sigma(h_t'(w')).$ \Comment{Eq. (1)}
    \State $Q_w\gets Q_w\cup\{w',p'\}$.
\EndFor
\For{e in epochs}
    \State Sample mini-batch $\{x_t^i\} \in \mathcal{X}_t$
    \State Draw $t_w\sim\mathcal{T}_w$. Transform $x_w=t_w(x_t^i)$.
    \State Draw $t_s\sim\mathcal{T}_s$. Transform $x_q=t_s(x_t^i)$.
    \State Draw $t_s'\sim\mathcal{T}_s$. Transform $x_k=t_s'(x_t^i)$.
    \State $w=f_t(x_w)$. Compute $\hat{y}$ according to Eq. (3)(4)
    \State $q=f_t(x_q), k=f_t'(x_k)$. Compute $L_t^{ctr}$ according to Eq. (5)(6)
    \State $p_q=\sigma(g_t(x_q))$. Compute $L_t^{ce}$ and $L_t^{div}$ according to Eq. (7)(8)(9)
    \State $L_t=\gamma_1 L_t^{ce}+\gamma_2 L_t^{ctr}+\gamma_3 L_t^{div}$. \Comment{Eq. (10)}
    \State $\theta_t\gets \theta_t - \eta\triangledown L_t$ \Comment{Update target model}
    \State $\theta_t'\gets m\theta_t'+(1-m)\theta_t$ \Comment{Update momentum model}
    \State Compute $w', p'$ according to Eq. (1). Update $Q_w$.
    \State Update $Q_s$ with $k$.
\EndFor
\Ensure Adapted target model $g_t(\cdot)$
\end{algorithmic}
\end{algorithm}

\section{More Experiment Details}
\subsection{Data augmentation}
The distributions of weak, strong data augmentation $T_w, T_s$ we used are detailed in \cref{alg:dataaugmentation}. The weak distribution consists of common resizing, random cropping, and random horizontal flip, whereas the strong distribution follows the design in MoCo v2 \cite{chen2020improved}.

\begin{algorithm}
\label{alg:dataaugmentation}
\caption{Data augmentations in Pytorch style}
\begin{pynoborder}[frame=none]
from torchvision.transforms import *

weak_transform = Compose(
    [
        Resize((256, 256)),
        RandomCrop(224),
        RandomHorizontalFlip(),
        ToTensor(),
        Normalize(
            mean=[0.485, 0.456, 0.406],
            std=[0.229, 0.224, 0.225]
        ),
    ]
)

strong_transform = transforms.Compose(
    [
        RandomResizedCrop(224, scale=(0.2, 1.0)),
        RandomApply(
            [ColorJitter(0.4, 0.4, 0.4, 0.1)],
            p=0.8,  # not strengthened
        ),
        RandomGrayscale(p=0.2),
        RandomApply(
            [GaussianBlur([0.1, 2.0])], p=0.5
        ),
        RandomHorizontalFlip(),
        ToTensor(),
        Normalize(
            mean=[0.485, 0.456, 0.406],
            std=[0.229, 0.224, 0.225]
        ),
    ]
)

\end{pynoborder}
\end{algorithm}

\subsection{DomainNet-126}
We follow \cite{saito2019semi} to use 126 classes from the original 345 classes from DomainNet, which we referred to as DomainNet-126. The classes (with class ids) that we ended up using are shown in \cref{tab:dn126classes}. We also follow the item lists from the official website of DomainNet \cite{peng2019moment} \url{http://ai.bu.edu/M3SDA/} to remove the duplicated images, by finding the difference between the cleaned and old dataset versions. The resulted number of images for 4 domains are listed in \cref{tab:dnimagecount}. We will include the resulted item lists along with our code release as a reference.

\begin{table}[]
\centering
\caption{Classes used for DomainNet-126 with associated class ids.}
\resizebox{0.99\columnwidth}{!}{
\begin{tabular}{c|c|c|c|c|c}
\toprule
id & class             & id & class      & id  & class                       \\
\midrule
0  & aircraft\_carrier & 42 & dolphin    & 84  & penguin                     \\
1  & alarm\_clock      & 43 & dragon     & 85  & pig                         \\
2  & ant               & 44 & drums      & 86  & pillow                      \\
3  & anvil             & 45 & duck       & 87  & pineapple                   \\
4  & asparagus         & 46 & dumbbell   & 88  & potato                      \\
5  & axe               & 47 & elephant   & 89  & power\_outlet               \\
6  & banana            & 48 & eyeglasses & 90  & purse                       \\
7  & basket            & 49 & feather    & 91  & rabbit                      \\
8  & bathtub           & 50 & fence      & 92  & raccoon                     \\
9  & bear              & 51 & fish       & 93  & rhinoceros                  \\
10 & bee               & 52 & flamingo   & 94  & rifle                       \\
11 & bird              & 53 & flower     & 95  & saxophone                   \\
12 & blackberry        & 54 & foot       & 96  & screwdriver                 \\
13 & blueberry         & 55 & fork       & 97  & sea\_turtle                 \\
14 & bottlecap         & 56 & frog       & 98  & see\_saw                    \\
15 & broccoli          & 57 & giraffe    & 99  & sheep                       \\
16 & bus               & 58 & goatee     & 100 & shoe                        \\
17 & butterfly         & 59 & grapes     & 101 & skateboard                  \\
18 & cactus            & 60 & guitar     & 102 & snake                       \\
19 & cake              & 61 & hammer     & 103 & speedboat                   \\
20 & calculator        & 62 & helicopter & 104 & spider                      \\
21 & camel             & 63 & helmet     & 105 & squirrel                    \\
22 & camera            & 64 & horse      & 106 & strawberry                  \\
23 & candle            & 65 & kangaroo   & 107 & streetlight                 \\
24 & cannon            & 66 & lantern    & 108 & string\_bean                \\
25 & canoe             & 67 & laptop     & 109 & submarine                   \\
26 & carrot            & 68 & leaf       & 110 & swan                        \\
27 & castle            & 69 & lion       & 111 & table                       \\
28 & cat               & 70 & lipstick   & 112 & teapot                      \\
29 & ceiling\_fan      & 71 & lobster    & 113 & teddy-bear                  \\
30 & cello             & 72 & microphone & 114 & television                  \\
31 & cell\_phone       & 73 & monkey     & 115 & The\_Eiffel\_Tower          \\
32 & chair             & 74 & mosquito   & 116 & The\_Great\_Wall\_of\_China \\
33 & chandelier        & 75 & mouse      & 117 & tiger                       \\
34 & coffee\_cup       & 76 & mug        & 118 & toe                         \\
35 & compass           & 77 & mushroom   & 119 & train                       \\
36 & computer          & 78 & onion      & 120 & truck                       \\
37 & cow               & 79 & panda      & 121 & umbrella                    \\
38 & crab              & 80 & peanut     & 122 & vase                        \\
39 & crocodile         & 81 & pear       & 123 & watermelon                  \\
40 & cruise\_ship      & 82 & peas       & 124 & whale                       \\
41 & dog               & 83 & pencil     & 125 & zebra      \\
\bottomrule
\end{tabular}
}
\label{tab:dn126classes}
\end{table}

\begin{table}[]
\centering
\caption{Image count of the 4 domains of the resulted DomainNet-126.}
\resizebox{0.4\columnwidth}{!}{
\begin{tabular}{cc}
\toprule
Domain   & Image count \\
\midrule
Real     & 69622       \\
Sketch   & 24147       \\
Clipart  & 18523       \\
Painting & 30042       \\
\bottomrule
\end{tabular}
}
\label{tab:dnimagecount}
\end{table}

{\small
\bibliographystyle{ieee_fullname}
\bibliography{egbib}
}
